%% file: Formatting-Instructions-LaTeX-2021.tex
\def\year{2021}\relax
\documentclass[letterpaper]{article} 
\usepackage{aaai21}  
\usepackage{times}  
\usepackage{helvet} 
\usepackage{courier}  
\usepackage[hyphens]{url}  
\usepackage{graphicx} 
\usepackage{natbib}  
\usepackage{caption} 
\usepackage{booktabs}
\usepackage{amsmath}
\usepackage{amssymb}

\usepackage[switch]{lineno}
\usepackage{algorithm}
\usepackage[noend]{algorithmic}
\usepackage{makecell}
\usepackage{enumitem}

\newcolumntype{P}[1]{>{\centering\arraybackslash}p{#1}}

\title{
Attribute-Guided Adversarial Training for Robustness to Natural Perturbations
}

\author{
    Tejas Gokhale\textsuperscript{1}\footnote{Work performed during internship at LLNL.}, 
    Rushil Anirudh\textsuperscript{2}, 
    Bhavya Kailkhura\textsuperscript{2}, 
    Jayaraman J. Thiagarajan\textsuperscript{2}, \\
    Chitta Baral\textsuperscript{1}, 
    Yezhou Yang\textsuperscript{1}\\}
\affiliations{
\textsuperscript{1} Arizona State University, 
\textsuperscript{2} Lawrence Livermore National Laboratory\\
\{tgokhale, chitta, yz.yang\}@asu.edu, \{anirudh1, kailkhura1, jjayaram\}@llnl.gov
}
\begin{document}

\maketitle
\begin{abstract}

While existing work in robust deep learning has focused on small pixel-level norm-based perturbations, this may not account for perturbations encountered in several real-world settings.
In many such cases although test data might not be available, broad specifications about the types of perturbations (such as an unknown degree of rotation) may be known.
We consider a setup where robustness is expected over an unseen test domain that is not i.i.d.\ but deviates from the training domain.
While this deviation may not be exactly known, its broad characterization is specified \textit{a} priori, in terms of attributes.
We propose an adversarial training approach which learns to generate new samples so as to maximize exposure of the classifier to the attributes-space, without having access to the data from the test domain.
Our adversarial training solves a min-max optimization problem, with the inner maximization generating adversarial perturbations,
and the outer minimization finding model parameters by optimizing the loss on adversarial perturbations generated from the inner maximization.
We demonstrate the applicability of our approach on three types of naturally occurring perturbations --- object-related shifts, geometric transformations, and common image corruptions.
Our approach enables deep neural networks to be robust against a wide range of naturally occurring perturbations.
We demonstrate the usefulness of the proposed approach by showing the robustness gains of deep neural networks trained using our adversarial training on MNIST, CIFAR-10, and a new variant of the CLEVR dataset.

\end{abstract}

\section{Introduction}
The goal of \textit{robust} machine learning models for tasks such as image classification is to make accurate predictions on \textit{unseen} samples.
The i.i.d.\ assumption is the simplest case in which unseen samples come from the same distribution as the training dataset.
However, in most real-world situations, this assumption breaks down and so do models trained under the i.i.d.\ paradigm~\cite{recht2018cifar, bulusu2020anomalous}.

Most work on adversarial robustness has focused on pixel-level $\ell_p$ norm-bounded perturbations such as additive noise~\shortcite{goodfellow2014explaining,sinha2018certifying,madry2018towards,raghunathan2018certified}.
While such perturbations allow the use of tractable mathematical formulations, in practice, they are not the only perturbations that might be encountered at test time.
For example, geometric transforms such as rotation, translation, or scaling of images, that are commonly encountered in the real world are not accounted for by pixel-wise $\ell_p$ bounded perturbations.

As such, images are parameterized by several unique attributes ranging from low-level information responsible for image formation like lighting, camera angle and resolution; to high-level semantic information like changes in background, size, shape, or color of objects in a scene.
Perturbations along many of these attributes are irrelevant to tasks like image classification and are thus ``semantics-preserving" perturbations.
For instance, translating a digit inside an image in a digit classification task, or manipulating the shape of an object in a color classification task, will not result in a change in the true class-label.
Yet, perturbations along these attributes are likely to cause models to fail when they are changed intentionally or otherwise \cite{xiao2020noise,joshi2019semantic,liu2018beyond}.
Shifts in such ``nuisance attributes'' typically result in large $\ell_p$ perturbations, posing significant challenges for existing pixel-level perturbation models.
On the other hand, it is impractical to sample the entire attribute space effectively in order to guard against potential failures at test time.

In this work, we propose a robust modeling technique for image classification problems, which learns to generate new samples so as to maximize the exposure of the classifier to variations in the attribute space.
Our approach falls under the broad category of adversarial training~\cite{madry2017towards}, and utilizes a min-max optimization setup, wherein the inner maximization step generates adversarial attribute perturbations while the outer minimization step identifies model parameters that reduce the task-specific loss (e.g., categorical cross entropy) under these perturbations.
We find that the attribute-based specification produces models that can more effectively handle challenging real-world distribution shifts than standard $\ell_p$ norm-bounded perturbations~\cite{qiao2020learning}.
Furthermore, our proposed approach is flexible to support a wide-range of attribute specifications, which we demonstrate with three different use-cases:
\begin{enumerate}[nosep,noitemsep,leftmargin=2em]
    \item Object-level shifts from a conditional GAN for adversarial training on a new variant of the CLEVR dataset;
    \item Geometric transformations implemented using a spatial transformer for MNIST data; and
    \item Synthetic image corruptions on CIFAR-10 data.
\end{enumerate}
Our contributions can be summarized as follows:
\begin{itemize}[nosep,noitemsep,leftmargin=2em]
    \item We consider the problem of robustness under a set of specified attributes, that go beyond typically considered $\ell_p$ robustness in the pixel space.
    \item We present Attribute-Guided Adversarial Training (AGAT), a robust modeling technique that solves a min-max optimization problem and learns to explore the attribute space and to manipulate images in novel ways without access to any test samples.
    \item We create a new benchmark called ``CLEVR-Singles''
    to evaluate robustness to semantic shifts. The dataset consists of images with a single block having variable colors, shapes, sizes, materials, and position.
    \item We demonstrate the efficacy of our method on three classes of semantics-preserving perturbations: object-level shifts, geometric transformations, and common image corruptions.
    \item Our method outperforms competitive baselines on three robustness benchmarks: CLEVR-Singles, MNIST-RTS, and CIFAR10-C.
\end{itemize}

\section{Related Work}
\input{related_work}

\input{method}

\section{Experiments}
\input{experiments}

\section{Conclusion}
In this paper, we propose a new adversarial training strategy for robustness against large perturbations that are common in practical settings.
Our adversarial training algorithm perturbs the attribute space to synthesize new images instead of pixel-level perturbations which are common to the robustness literature.
The new CLEVR-Singles dataset that we have created can be used in future work for studying robustness to semantic shifts.
We extensively evaluate AGAT training on three benchmarks and achieve state-of-the-art performance.
We empirically show that AGAT is applicable to a three types of naturally occurring perturbations, and can be used with different classes of surrogate functions.
AGAT can potentially be applied to a broad range of robustness problems not limited to classification.

\section*{Acknowledgments} 
This work was performed under the auspices of the U.S. Department of Energy by the Lawrence Livermore National Laboratory under Contract No. DE-AC52-07NA27344, Lawrence Livermore National Security, LLC. This document was prepared as an account of the work sponsored by an agency of the United States Government. Neither the United States Government nor Lawrence Livermore National Security, LLC, nor any of their employees makes any warranty, expressed or implied, or assumes any legal liability or responsibility for the accuracy, completeness, or usefulness of any information, apparatus, product, or process disclosed, or represents that its use would not infringe privately owned rights. Reference herein to any specific commercial product, process, or service by trade name, trademark, manufacturer, or otherwise does not necessarily constitute or imply its endorsement, recommendation, or favoring by the United States Government or Lawrence Livermore National Security, LLC. The views and opinions of the authors expressed herein do not necessarily state or reflect those of the United States Government or Lawrence Livermore National Security, LLC, and shall not be used for advertising or product endorsement purposes. This work was supported by LLNL Laboratory Directed Research and Development project 20-ER-014 and released with LLNL tracking number LLNL-JRNL-814425.

\section*{Broader Impact}
The concept of robustness is critical when it comes to deploying machine learning systems in practical settings where input signals may undergo perturbations due to weather (such as fog, smog, rain), digital corruptions in transmission, or changes in camera inclinations causing geometric transformations or artifacts such as defocusing, or motion blur.
Our method for developing robust classifiers is broadly applicable if such classes of perturbations are known \textit{a priori}.

Robustness research is also crucial for avoiding or removing unintended biases that may percolate from the training data into the classification model.
Recent studies~\cite{bolukbasi2016man,zhao2017men,hendricks2018women} have shown that models trained on biased data can in fact amplify this bias when performing inference on test samples.
We believe that work in the lines of AGAT could be potentially used for mitigating social biases due to biased training data, such as gender or racial biases.

\bibliography{ref}

\appendix 
\input{appendix_arxiv}
\end{document}

%% file: related_work.tex
Most existing work on robustness deals with the problem of finding $\ell_p$ perturbations which focus on additive noise, with tractable mathematical guarantees of performance when test data falls within an $\epsilon$-ball of the training distribution~\cite{goodfellow2014explaining,sinha2018certifying,madry2018towards,raghunathan2018certified}.
Such perturbation are typically \emph{imperceptible} to the human eye. 
As a result, there is an increasing interest in addressing challenges that arise from natural corruptions or perturbations \cite{hendrycks2018benchmarking} that are \emph{perceptible} shifts in the data, more likely to be encountered in the real world.
For example,~\cite{liu2018beyond} use a differentiable renderer to design adversarial perturbations sensitive to semantic concepts like lighting and geometry in a scene;~\cite{joshi2019semantic} design perturbations only along certain pre-specified attributes by optimizing over the range-space of a conditional generator. 
Our work focuses on building robust models against semantic, or more generally attribute guided concepts that may or may not exist in the training distribution, using a surrogate function. 

$\ell_p$-norm based robustness methods make no assumptions about the test distribution, except that the methods are guaranteed to be robust only inside the $\epsilon$-ball of the training distribution~\cite{volpi2018generalizing,qiao2020learning}.
Some recent approaches extend this notion to assume some access to data from the test distribution such as TTT~\cite{sun2020test} that achieves robustness for a test example by minimizing the cost of an auxiliary task for each test sample; and~\cite{wong2020learning} learn a CVAE using possible corruptions one might encounter, to then guarantee robustness of a classifier within the learned perturbation set.
For comparison, our method assumes access to no data from the test distribution, but only knowledge of a specification, which is the intended functionality of the system specified in human-understandable attributes. Under this challenging set-up, we show our method still outperforms existing robustness techniques on popular and standard benchmarks.

%% file: method.tex
\section{Problem Setup}

\begin{figure}
    \centering
    \includegraphics[width=\linewidth]{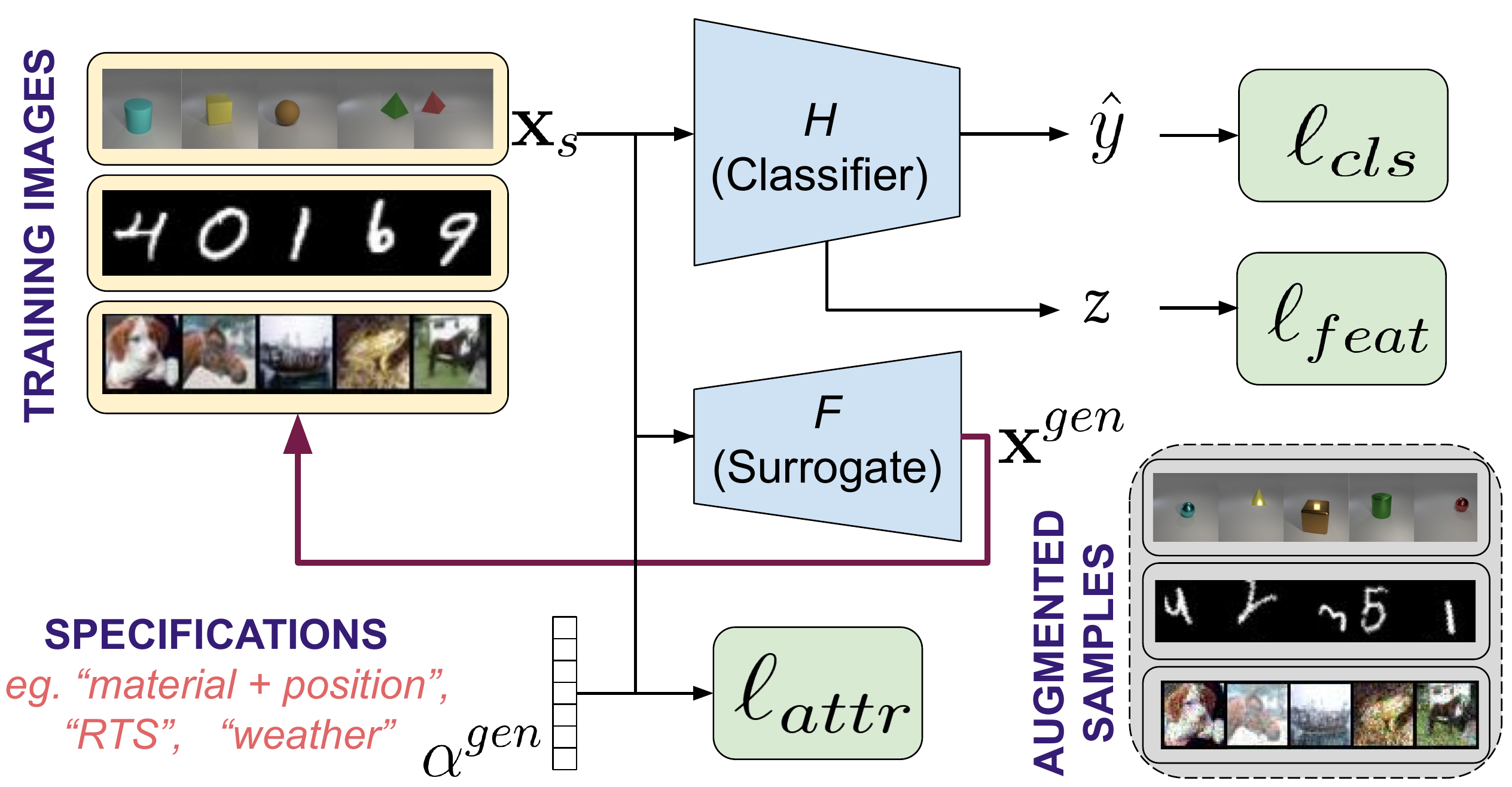}
    \caption{Overview of the problem setup and our attribute-guided adversarial training method.}
    \label{fig:problem}
\end{figure}

We begin by defining the classifier parameterized by a set of neural network weights, $\theta$, as
${H}_{\theta}:
\mathcal{X}_s\mapsto \mathcal{Y}$, where $\mathcal{X}_s$ denotes the space of the observed image data (or source) and $\mathcal{Y}$ denotes the label space for the task of interest.

\subsubsection{Robustness to natural perturbations}
Our goal is to train an $H_{\theta}$ that is robust to \emph{natural} perturbations, which are typically larger in magnitude than the \emph{imperceptible} $\ell_p$-bounded pixel-space perturbations, considered in the literature.

We consider a broad range of natural semantics-preserving perturbations that will not affect the predictions for the task under consideration --
    (a) \textbf{Object-level shifts}, where attributes of the object are
    manipulated so as to considerably change the appearance of the object, without changing the task label; such as changing the shape or size of an object in a color classification task.
    (b) \textbf{Geometric transformations}, where the test image may be scaled, rotated, and shifted in arbitrary ways; and
    (c) \textbf{Common image corruptions}, which may occur in the real world like fog, image compression artifacts, blurs, and other forms of noise.

Most of these perturbations do not naturally fall within small $\ell_p$-norm ball deviations ($|\mathbf{x}-\tilde{\mathbf{x}}|_p\leq \epsilon$), for which most existing robustness methods are designed, and are bound to fail when the classifier encounters such data in the wild.
However, making $\epsilon$ arbitrarily large in robustness formulations does not work in practice, since the image quality degrades significantly. 
Hence, we propose a new framework to design models that are robust to such natural perturbations.

\section{Attribute Guided Adversarial Training}

Let us denote an image by $\mathbf{x}_\alpha$ parameterized by a set of attributes $\alpha$ related to image formation (lighting, viewing angle, position) as well as abstract semantic information (color, shape, size, etc.). 
Manipulating images with new combinations of attributes that are not seen in the training set, requires access to the underlying physical generative processes,  which is unrealistic. We do not assume direct access to such deterministic mechanisms.

Our goal is to train classifiers robust to natural perturbations along 
attributes in $\alpha$ that are specified \emph{a priori}. Inspired by recent developments in robust optimization and adversarial training~\cite{madry2018towards}, we consider the following worst-case problem around $N$ attributes of the training data:
\begin{equation}
    \small
    \min_{\theta\in \Theta}\sum_{i=1}^N\max_{|\hat{\alpha}_i-\alpha_i|\leq \epsilon} \ell(\theta;(\mathbf{x}_{\hat{\alpha}_i},y_i)),
    \label{eq:adv_training_worst_case}
\end{equation}
where $\ell(\cdot)$ is the cross-entropy loss.

The solution to worst-case optimization in Equation~\ref{eq:adv_training_worst_case} guarantees good performance against test data that is distance $\epsilon$ away from the training data in the attribute space. 
In other words, we expect the model learned using \eqref{eq:adv_training_worst_case} to be robust against $\epsilon$-bounded natural perturbations.
Interestingly, as we will empirically show later, models learnt using \eqref{eq:adv_training_worst_case} perform better than existing pixel-level techniques even on $\ell_p$-bounded imperceptible perturbations.

Although the structure of the attribute-guided adversarial training problem may look similar to standard adversarial training, we explain next why solving Eq~\eqref{eq:adv_training_worst_case} is significantly more challenging and requires us to make several algorithmic innovations. 
Note that \eqref{eq:adv_training_worst_case} solves a min-max optimization problem, with the inner maximization generating natural perturbations by maximizing the classification loss over attribute space, and the outer minimization finding model parameters by minimizing the loss on natural perturbations of the training data generated from the inner maximization. The success of this method crucially relies on solving the inner optimization problem. 
Motivated by the standard adversarial training, one might be temped to approximately solve the re-parameterized inner optimization problem
\begin{equation}
    \small
    \underset{|{\delta}|_p\leq \epsilon}{\max} l(\theta;(\mathbf{x}_{{\alpha}_i+{\delta}},y_i)),
\end{equation}
and generate the natural perturbations
$\mathbf{x}_{\hat{\alpha}_i}^*$
using projected gradient descent (PGD) as:
\begin{equation}
    \small
    {\delta_i}^* := \mathcal{P}_{\epsilon} (\delta_i-\lambda \nabla_{\delta} l(\theta;(\mathbf{x}_{{\alpha}_i+{\delta}},y_i))),
\end{equation}
where $\lambda$ is the gradient step and $\mathcal{P}_{\epsilon}$ is projection on $l_p$ ball of radius $\epsilon$. However, there are two fundamental issues with this approach making it infeasible in practice: first, we cannot compute gradients as we do not have access to the attribute space;
and second, we do not have access to the true generative mechanism conditioned on the attributes.

\subsection{Proposed Approach}
\subsubsection{Surrogate Functions}
We propose to use differentiable surrogate functions parameterized by attributes to overcome the limitation described above. 
In other words, we have 
$\mathbf{x}_{\alpha+\delta} \approx F_{\delta}(\mathbf{x}_{\alpha})$, 
where $F_{\delta}$ is differentiable.
Typically, exact perturbations 
$\mathbf{x}_{\alpha+\delta} = F_{\delta}(\mathbf{x}_{\alpha})$
can be performed for PGD attacks or other $\ell_p$ norm bounded attacks.
However, in our case accessing the true generative process to manipulate images along $\alpha$ is not feasible. 
For example, we cannot rely on deterministic functions to manipulate semantic features in the image like size, shape or texture of an object.
As a result, we resort to using \emph{approximate} image manipulators in the form of surrogate functions which act as proxies to the true generative process.
Depending on the type of attributes against which we wish to train for robustness, the surrogate function can take different forms:
\begin{itemize}[nosep,noitemsep,leftmargin=2em]
    \item generative editing models for semantic perturbation that is learned from the training data itself, 
    \item analytical functions for geometric transformations in the form of spatial transformers (STNs), or 
    \item an analytical approximation (or tractable upper bound) of the natural perturbation space.
\end{itemize}
For example, if we want a classifier robust to unknown affine transforms then $F$ is the spatial transformer layer parameterized by $\alpha$ which now represents $6$ parameters controlling rotation, scale, and shift of the image.

Note that we do not assume access to any additional data other than the clean training dataset $\mathcal{X}_s$, and specification of the class of functions against which robustness is desired. While such surrogate functions only approximate the natural perturbations, we show that they are sufficient for enabling us to make classifiers more robust to natural perturbations.

\subsubsection{Iterative Training Procedure}
Having access to attribute parameterized surrogate function, we aim to solve \eqref{eq:adv_training_worst_case}. Note, the success of the adversarial training is dependent on the quality of the generated perturbations. Thus, we aim to generate natural perturbations that have a larger coverage over the specified attribute space than the training samples images $\mathbf{x}_s$. Consider the classifier $H_\theta$ which outputs the predicted class $\hat{y}$ and intermediate features $z$, let the surrogate function ${F}$ be parameterized by the attribute vector $\alpha$. We propose an iterative training procedure called Attribute-Guided Adversarial Training (\textbf{AGAT}) detailed in Algorithm~\ref{alg:adv_algo}. Our algorithm has two objectives: to minimize the classification loss over input images and to maximize the divergence between the training samples and generated perturbations. Thus, the key idea here is to explore novel and hard images that only \emph{vary along the specified attributes}. To achieve this, we impose a constraint that maximizes the distance between features of dataset images and perturbed images. Additionally, since we would also like to explore new regions in the attribute space, we impose a similar constraint on the attributes of perturbed images.

We express this constraint as the loss function given by:
\begin{equation}
    \small
    \begin{split}
        \ell_{const} = \lambda_1\ell_{feat} + \lambda_2\ell_{attr},~~\lambda_1, \lambda_2 \in (0,1)\\
        \text{where~~ }
        \ell_{feat} = ||\mathbf{z} - \mathbf{z^{gen}}||_2^2,
        \text{~and~~}
        \ell_{attr} = ||\alpha - \alpha^{gen}||_2^2
    \end{split}
    \label{eq:loss_constraint}
\end{equation}

To ensure that the generated images belong to the same class as the input image, we combine classification loss with respect to the ground truth label $\mathbf{y}$ with consistency regularization with respect to the predicted label $\mathbf{\hat{y}}$ of $\mathbf{x}$.
\begin{equation}
    \small
    \ell_{cls} = \ell_{BCE}(\mathbf{y}, \mathbf{y^{gen}}) + \ell_{BCE}(\mathbf{\hat{y}}, \mathbf{y^{gen}})
    \label{eq:loss_distill}
\end{equation}

\noindent The overall loss function is computed as the Lagrangian:
\begin{equation}
    \small
    \ell_{AGAT} = \ell_{cls} - \beta\cdot\ell_{const}
    \label{eq:loss_overall}
\end{equation}

Intuitively, $\ell_{cls}$ encourages the augmented images to belong to the same class-label as the input image, while the constraint $\ell_{const}$ encourages the adversarial learning algorithm to perturb the image features as well as the attributes away from the input features and attributes.
\input{algo}
We first pre-train the classifier only on the source samples $\mathbf{x}_s$ for $N_{pre}$ epochs.
Then, we initiate our augmentation process.
To generate new samples, we minimize Equation~\ref{eq:loss_overall} and update the attribute vector for $M$ update steps as:
\begin{equation}
    \small 
    \alpha^{gen} \gets \alpha^{gen} - \mu\triangledown \ell_{AGAT}.
\end{equation}
Finally, synthetic images are generated using the surrogate function $\mathbf{x}^{gen} \gets F(\mathbf{x}, \alpha^{gen})$, and appended to the training data.
This adversarial data augmentation is performed after every $N_{aug}$ epochs during which $T_{aug}$ images are generated.
The total number of augmented samples is expressed as a percentage of the number of training samples so as to allow fair comparison across datasets and types of perturbations.
The pseudocode for AGAT is shown in Algorithm~\ref{alg:adv_algo}.

The distinguishing factor for \textbf{AGAT} is that we perturb the attribute space and use surrogate functions to synthesize images, while previous adversarial augmentation protocols such as M-ADA~\cite{qiao2020learning} and GUD~\cite{volpi2018generalizing} perturb only in the pixel-space, thus being restricted to $\ell_p$ perturbations.
It is important to note that our method is agnostic to the choice of surrogate functions, which can take the form of additive noise, affine transformation in pixel-space, or conditional generative adversarial networks~\cite{mirza2014conditional} trained to transform an input image according to an input attribute vector.

%% file: algo.tex
\begin{algorithm}[t]
    \caption{Attribute-Guided Adversarial Training}
    \label{alg:adv_algo}
    \small
    \begin{algorithmic}[0]
        \STATE \textbf{Input:} Source dataset $\mathcal{D}_S=\{\mathbf{x}_t,y_t\}_{i=t}^T$
        \STATE \textbf{Output:} learned weights $\theta$ 
    \end{algorithmic}
    \begin{algorithmic}[1]
        \STATE \textbf{Initialize:} $\theta\gets\theta_0$, $\mathcal{D}^{aug}_S\gets\mathcal{D}_S$
        \FOR{$n=1\dots N_{epochs}$}
            \IF{$n < N_{pre}$}
                \FOR{$t=1:T$}
                    \STATE $\theta \gets \theta - \eta\triangledown\ell_{cls}(\theta;(\mathbf{x}_t, y_t))$
                \ENDFOR
            \ELSE
                \IF{$n~\text{mod }N_{aug} = 0$}
                    \FOR{$t=1\ldots T_{aug}$}
                        \STATE sample ${(\mathbf{x}_t, y_t)}_{t=1}^{T_{aug}}$ from $\mathcal{D}_S$
                        \STATE $z_t, \hat{y}_t = H(\mathbf{x_t})$
                        \STATE \textbf{Initialize:} $\alpha_t^{gen}$
                        \FOR{$i=1\ldots M$}
                            \STATE $z^{gen}_t, \hat{y}^{gen}_t = H(\mathbf{x}_t, \alpha_t^{gen})$
                            \STATE $\mathbf{x}_t^{gen} \gets f(\mathbf{x}_t, \alpha_t^{gen})$
                            \STATE $\alpha_t^{gen} \gets \alpha_t^{gen} - \mu\triangledown (\cdot\ell_{cls}-\beta\cdot \ell_{cons})$
                        
                        \ENDFOR
                        
                        \STATE $\mathcal{D}_S^{aug} \gets \mathcal{D}_S^{aug} \cup \mathbf{x}_t^{gen}$
                    \ENDFOR
                \ELSE
                    \FOR{$(\mathbf{x}_t, y_t) \in \mathcal{D}^{aug}_S$}
                         \STATE $\theta \gets \theta - \eta\triangledown\ell(\theta;(\mathbf{x}_t, y_t))$
                    \ENDFOR
                \ENDIF
            \ENDIF
        \ENDFOR
    \end{algorithmic}
    
\end{algorithm}
%

%% file: experiments.tex
In this section, we introduce the three types of robustness specifications that we experiment with, along with details about the datasets, baselines, and metrics used for each.

\begin{figure}[t]
    \centering
    \includegraphics[width=0.99\linewidth]{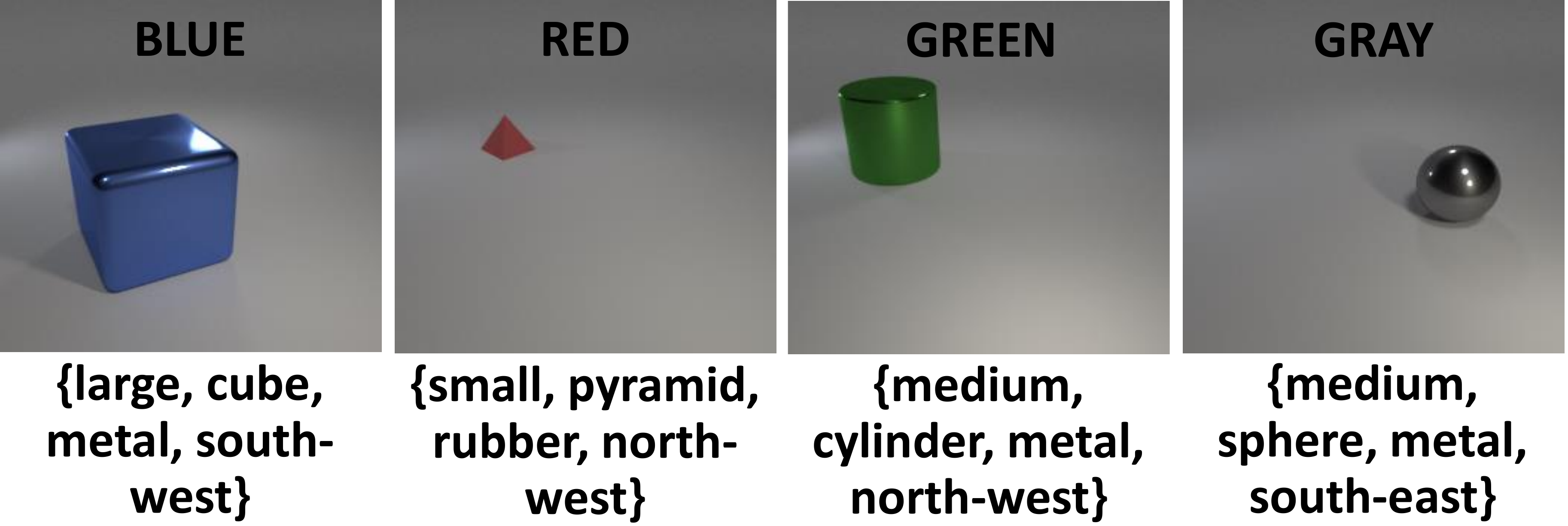}
    \caption{Examples of images from CLEVR-Singles and the color labels and (size, shape, material, position) attributes.}
    \label{fig:clevr_singles}
\end{figure}

\input{tables/clevr_splits}

\begin{figure}[t]
    \centering
    \includegraphics[width=0.75\linewidth]{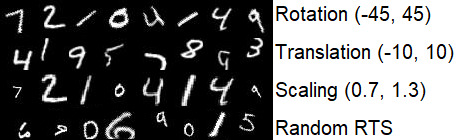}
    \caption{
    RTS-perturbed MNIST images.
    }
    \label{fig:mnist_rts}
\end{figure}

\subsection{Semantic Object-Level Perturbations}
One class of real-world perturbations is when properties or attributes of images or objects in images change at test time.
These changes do not affect the classification label, but significantly change the appearance of the image.
For instance, consider the task of color classification in objects of varied shapes and textures. 
Here, \textit{red metallic spheres} and \textit{red rubber cubes} both belong to the class label ``red", however may appear very different in their shapes and textures. 
Thus, if only \textit{red metallic cubes} are seen during training, conventional classifier predictions for test images consisting of \textit{red rubber cubes} can fail to generalize. 
Thus, although the class label is invariant to such semantic factors, robustness to perturbations along these factors is desirable.

    \subsubsection{Dataset:}

    To study the problem of such object-level shifts along semantic factors of an image in a controlled fashion, we create a new benchmark called CLEVR-Singles\footnote{
        Dataset: \text{https://github.com/tejas-gokhale/CLEVR-Singles}} 
    by modifying the data generation process from CLEVR~\cite{johnson2017clevr}.
    We create images of single objects having one of eight colors, and use color classification as our task in this paper.
    Each object has four variable attributes that do not affect the color class of the image; these are: \textit{shape} (cube, sphere, pyramid, or cylinder), \textit{size} (small, medium, or large), \textit{material} (rubber or metal), and \textit{position} (northwest, southwest, northeast, southeast).
    While the objects are generated at continuous $(X, Y, Z)$ world coordinates, we assign them a discrete position class for our experiments.
    Object-level perturbations can be made over these four attributes for our robustness experiments.
    In other words, it is known that one or more of \textit{\{shape, size, material, position\}} of the image may change at test-time without knowing the magnitude or combinations of the change.
    We split the dataset based on a combination of attributes as shown in Table~\ref{tab:clevr_splits}; for instance only certain combinations of size and position are observed in the training set, but robustness is expected from the color classifier on unknown combinations.

    \subsubsection{AttGAN as the Surrogate Function:}
    \begin{figure}
        \centering
        \includegraphics[width=\linewidth]{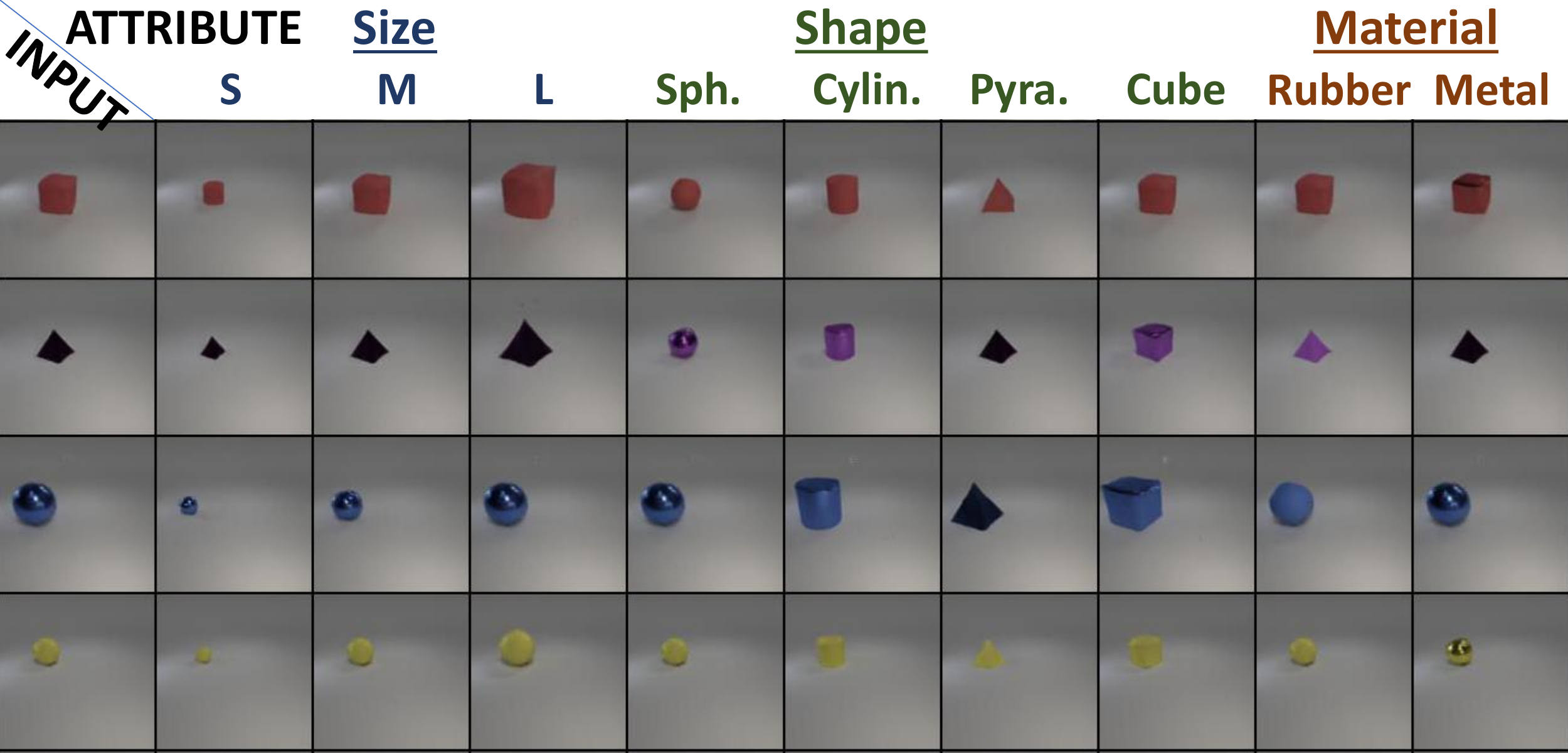}
        \caption{Images generated by AttGAN for the images in column 1, conditioned on attributes.
        }
        \label{fig:attgan}
    \end{figure}
    Conditional generative adversarial networks (cGANs) have been shown to perform exceptionally well on image-to-image translation in various domains~\cite{isola2017image,zhang2017stackgan,karras2019style}.
    AttGAN~\cite{he2019attgan} is one such conditional GAN which is trained to manipulate attributes of input face images.
    Thus given an image and a vector of desired attributes, AttGAN can manipulate the face image along the desired attribute dimensions.
    We leverage this powerful image manipulation technique as our surrogate function $\mathbf{x}^{gen} = F_{GAN}(\mathbf{x}, \alpha)$.
    Formally, we define the attribute vector to be a 13-dimensional binary hash-code with 1 and 0 indicating presence or absence of an attribute.
    For each experiment, we train the AttGAN on the training dataset outlined in Table~\ref{tab:clevr_splits} to generate 128x128 images, with a learning rate of $2\mathrm{e-}4$ for 100 epochs on a single 16GB GPU.
    Examples of images generated by AttGAN are shown in Figure~\ref{fig:attgan}, when manipulating certain attributes such as size, shape, and material of the object.

    \subsubsection{Baselines:}
    For the color classification task on CLEVR-Singles images, we compare against two pixel-level domain augmentation baselines: GUD~\cite{volpi2018generalizing} which performs adversarial data augmentation to generate fictitious target domains, and M-ADA~\cite{qiao2020learning} which uses a meta-learning framework to generate multiple domains of samples.
    We also report the performance of a classifier directly without any adversarial training as a naive baseline.
    The same classifier architecture is used for each baseline for fair comparison.
    All models are trained for $15$ epochs including pre-training epochs $N_{pre}=5$, batch-size $64$, and $M=15$ update steps for adversarial augmentation.
    The number of augmented samples $T_{aug}$ is $30\%$ of the original source data, and augmentation interval $N_{aug}$ is fixed at 2 epochs.
    For our model the coefficients in Equations~\ref{eq:loss_constraint}, and~\ref{eq:loss_distill} are: $\lambda_1=0.5, \lambda_2=0.5, \beta=0.25$.
    The learning rates $\eta, \mu$ for the classifier and adversarial augmentation are both 5e-5. 

    \subsubsection{Results:}
    \input{tables/clevr_results}
    The test classification accuracies for different splits are reported in Table~\ref{tab:clevr_results}. 
    We observe that our model is better than all baselines considered here, with a boost of $5$ percentage points in accuracy on the harder experiment along \textit{Material+Position}.

    \subsubsection{Analysis:}
    \begin{figure}[t]
        \centering
        \includegraphics[width=0.58\linewidth]{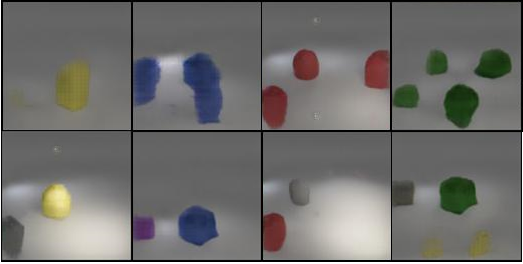}
        \caption{Visualization of the effect of weight $\beta$ of the constraint loss $\ell_{const}$ on the generated images. Row 2 has higher $\beta$ than Row 1. Illustration also shows that AttGAN is able to generate multiple objects (of same color for Row 1 and different colors for Row 2), though absent in training data.
        }
        \label{fig:clevr_novel_viz}
    \end{figure}

    In Figure~\ref{fig:clevr_novel_viz}, we show 8 different examples generated by AttGAN during adversarial training. We can see the effect of the coefficient $\beta$, from the constraint loss $\ell_{const}$ in eq~\eqref{eq:loss_overall}, in exploring the attribute space. An appropriately chosen value for $\beta$ encourages useful perturbations without violating the class-label consistency cost $\ell_{cls}$ as seen in the top row of Figure \ref{fig:clevr_novel_viz}. On the other hand, a higher $\beta$ would mean a higher weight for exploring the regions (or combinations) in attribute space not seen in training. In the bottom row we see that a high $\beta$ encourages novel attribute exploration at the cost of higher classification error as a result of generating objects with different colors within the same image.  It is noteworthy that AttGAN is able to generate images with multiple objects, even when it trained on images with only a single object, thus demonstrating its suitability to explore novel attributes using the proposed AGAT training.

\subsection{Geometric Transformations}
Another common class of perturbations is geometric transformations, i.e. a composition of rotation, translation, and scaling of an image. These perturbations are common since cameras may capture a scene from different orientations, distances, and inclinations. It is well known that standard image classifiers are not robust to these common perturbations~\cite{cohen2014transformation}.

    \subsubsection{Dataset:}
    We address this problem in the digit classification setting, with
    the training images from MNIST~\cite{lecun1998mnist}, and the test images that are perturbed along rotation-translation-scale (RTS), as shown in Figure~\ref{fig:mnist_rts}.
    We use the standard RTS setup~\cite{jaderberg2015spatial} with angle of rotation in $(-45, 45)^{\circ}$, translation in $(-10, 10)$ pixels in both directions, and a scale factor in the range $(0.7, 1.3)$.

    \subsubsection{Surrogate Function:}
    The attributes of interest, $\alpha$, consist of a $2\times3$ affine matrix that controls rotation, translation, and scale. To perform affine transformations on the image with a perturbed $\alpha$, we use Spatial Transformer Networks (STN)~\shortcite{jaderberg2015spatial} which allow differentiable spatial manipulation of input images in a convolutional neural network, such as RTS and or general warping. The perturbed images are generated as: $\mathbf{x}^{gen} = F_{STN}(\mathbf{x}_s, \alpha)$.

    \subsubsection{Baselines:}
    We compare the robustness performance to RTS perturbations with a naive baseline, denoted by (B), that is only trained on the standard MNIST dataset, and pixel-level perturbation methods MADA~\cite{qiao2020learning} and GUD~\cite{volpi2018generalizing}.
    Additionally, we also use the RTS perturbation sets generated by~\cite{wong2020learning} (PS) and use them as augmented training samples. 
    All models are trained for 12 epochs including pre-training epochs $N_{pre}=5$, with a batch-size $64$, and $M=10$ update steps for adversarial augmentation. The number of augmented samples $T_{aug}$ is $30\%$ of the original source data, and augmentation interval $N_{aug}$ is fixed at 10 epochs.
    Our model the coefficients in Equations~\ref{eq:loss_constraint}, and~\ref{eq:loss_distill} are: $\lambda_1=1, \lambda_2=1, \beta=5$.
    The learning rate $\eta$ for the classifier is $1\mathrm{e-}4$ and $\mu$ for the adversarial augmentation is $0.1$.

    \subsubsection{Results:}

    \input{tables/mnist_rts}
    We report digit classification accuracies on the target test set containing only rotations (R), only translations (T), only skew (S), as well as a random combination of RTS. Our model performs well on all four metrics, and beats the perturbation sets (PS) even though their augmentation model has access to RTS perturbations during training. In particular, we observe a significant improvement compared with MADA and GUD, in the robustness on the translation experiment, which is the hardest task among the three. 

    \subsubsection{Analysis}
     \begin{figure*}
        \centering
        \includegraphics[width=0.31\linewidth]{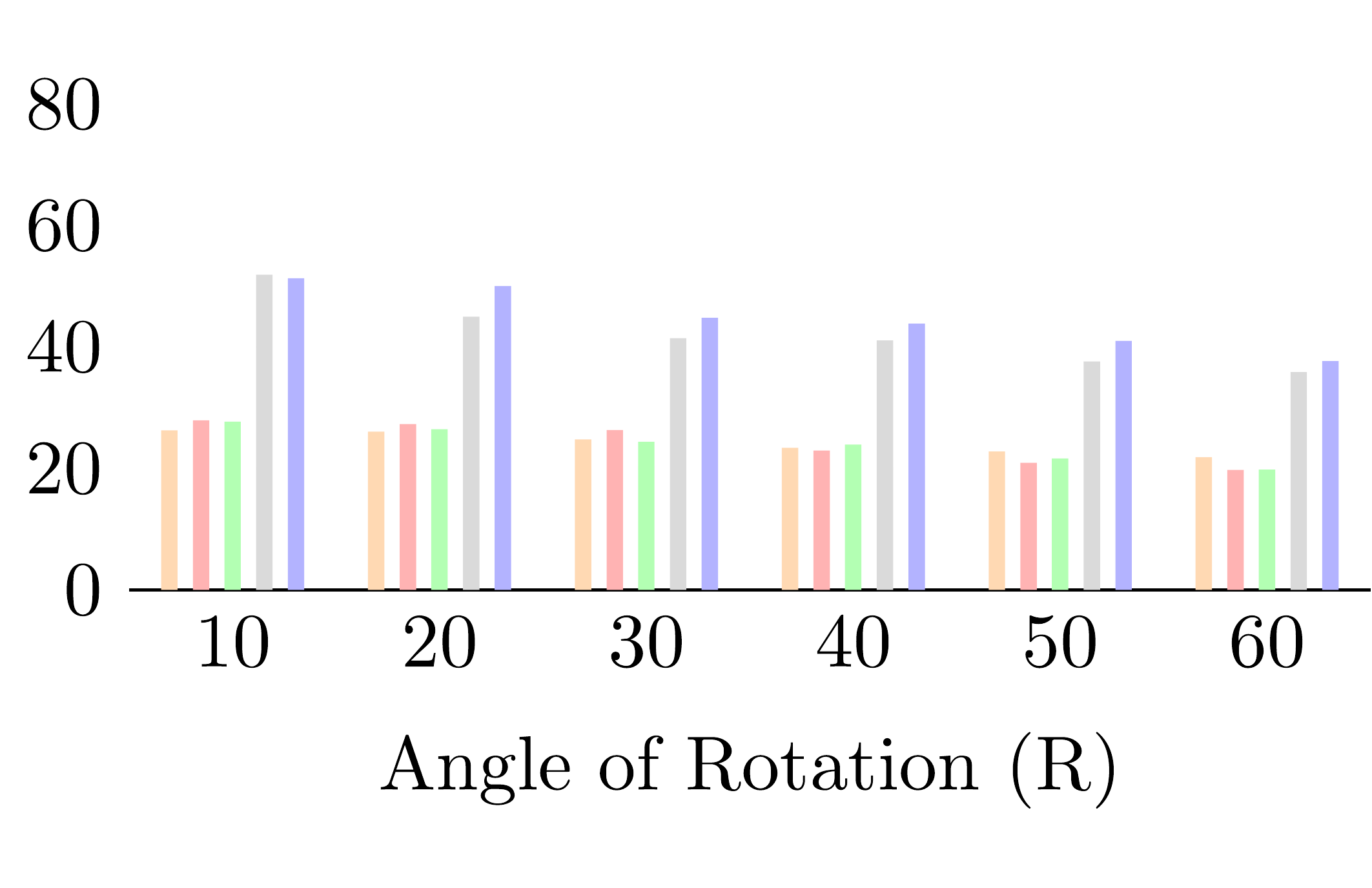}
        \includegraphics[width=0.31\linewidth]{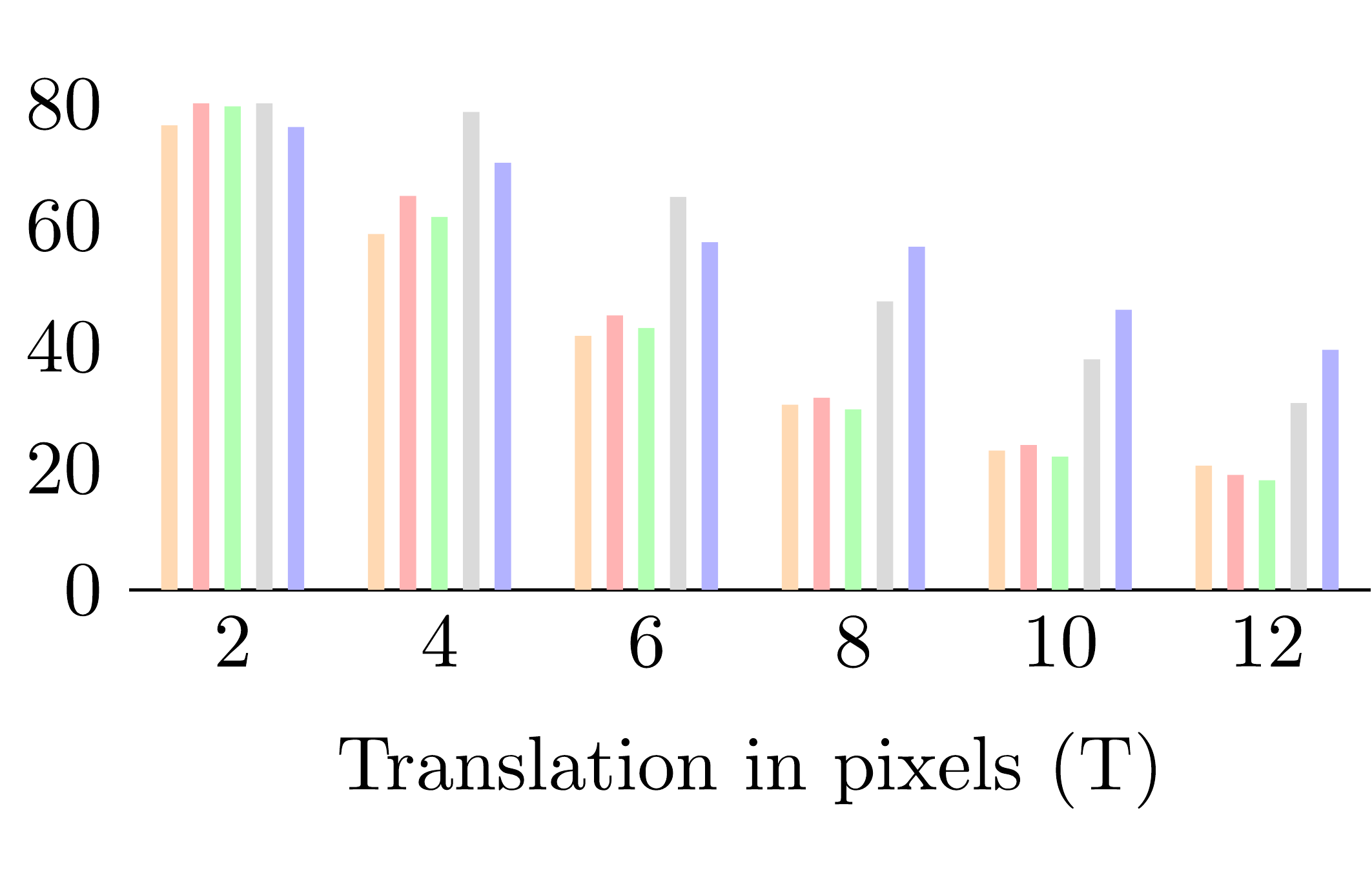}
        \includegraphics[width=0.31\linewidth]{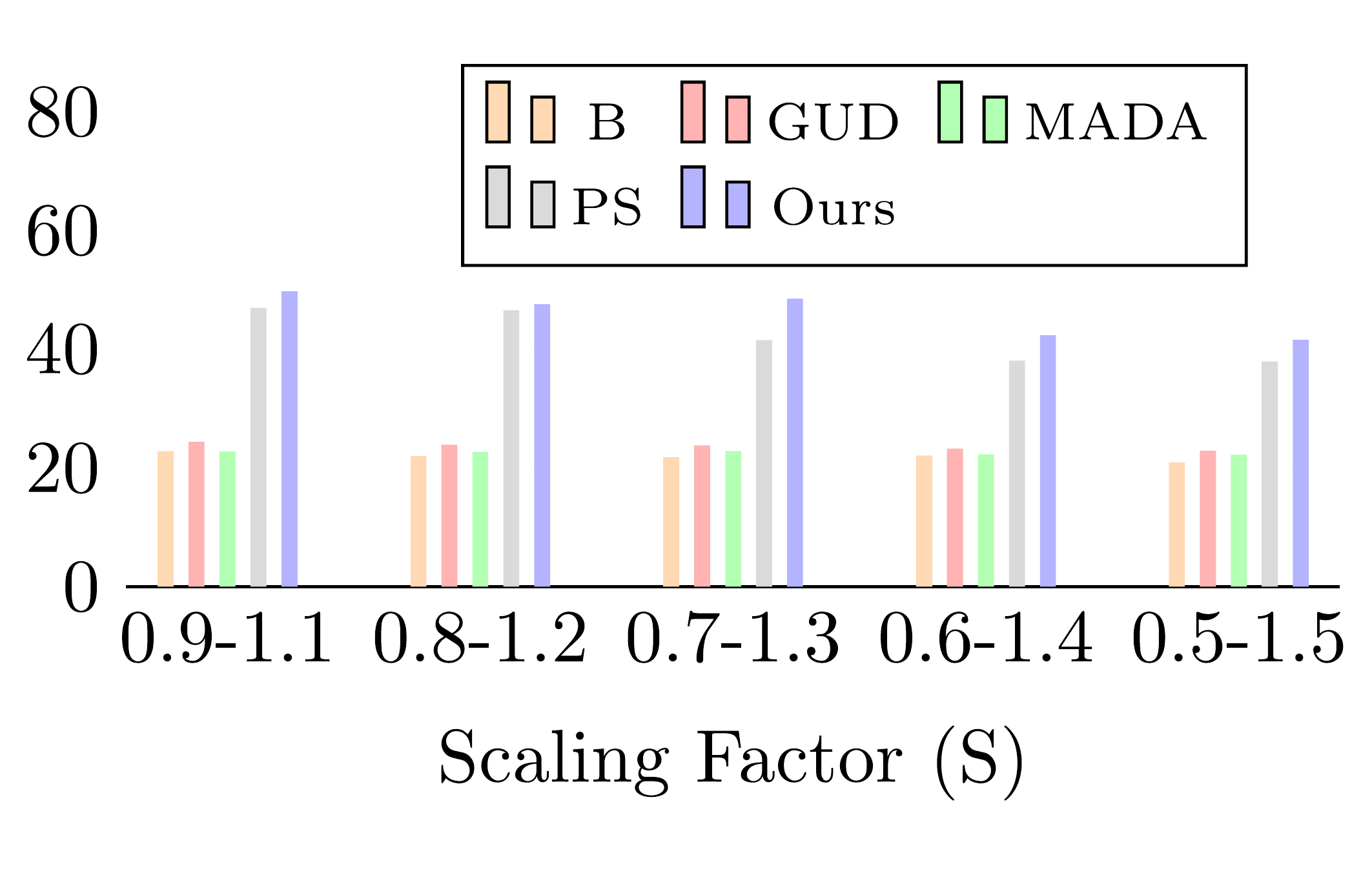}
        \caption{Comparison of random RTS accuracies when controlling each parameter to a max. value. Left: R, Center: T, Right: S}
    \end{figure*}
    \input{tables/mnist_ablation}
    The pixel-level perturbation methods still perform reasonably well on rotation and scale experiments in Table~\ref{tab:mnist_geom} because in each case the rotations/translations/scale are randomly sampled, resulting in several test examples that are very close to the training examples (with no RTS). In order to resolve this further, we study the performance by controlling the magnitude of R, T, and S in the test set. Figure~\ref{fig:mnist_rts} shows the bar-plots when the range of rotation is varied from $(-10, 10)$ to $(-60, 60)$, translation from $(-2, 2)$ to $(-12, 12)$ pixels, and scaling factor from $(0.9, 1.1)$ to $(0.5, 1.5)$.
    It can be observed that at higher severity of perturbation, our model (in blue) significantly outperforms all baselines.
    The model trained with Perturbations Sets~\shortcite{wong2020learning} (in gray) is competitive at lower severities.

    We also analyze the effect of the number of augmented samples ($T_{aug}$) expressed as a percentage of the size of the training data, while controlling for the augmentation interval $N_{aug}$. As expected, larger number of augmented samples improve robustness even higher than in table \ref{tab:mnist_geom} (which fixes number of additional augmented examples at $30\%$ for all baselines). 
    Larger augmentation intervals contribute positively at lower percentages of augmented samples.
    
    Finally, we perform an ablation study with and without the consistency regularization defined in Eq~\eqref{eq:loss_distill} and show that the regularization indeed helps improve performance.

\subsection{Common Image Corruptions}
Image corruptions are another common class of perturbations. These can occur due to image digitization artifacts, weather, camera calibration, and other sources of noise.

    \subsubsection{Dataset}
    The CIFAR10 dataset~\cite{krizhevsky2009learning}
    contains $50k$ training images belonging to $10$ classes.
    Recently, CIFAR10-C~\cite{hendrycks2018benchmarking} which contains image corruptions for CIFAR10 images, was proposed to benchmark robustness of image classifiers, with 4 major and  15 fine-grained categories of corruption: \textit{Weather} (fog, snow, frost), \textit{Blur} (zoom, defocus, glass, motion), \textit{Noise} (shot, impulse, Gaussian), and \textit{Digital} (JPEG, pixelation, elastic transform, brightness, contrast).
    There are five levels of severity of corruptions; we focus on the highest severity.

    \subsubsection{Surrogate Function:}
    We use a general surrogate function -- a composition of additive Gaussian noise and Gaussian blur filter parameterized by $\mathbf{\alpha} = \{\alpha_1, \alpha_2\}$:
    \begin{equation}
        \small
        \mathbf{x}^{gen} = \frac{1}{\sqrt{2\pi\alpha_1^2}}e^{-\frac{\mathbf{x}^2}{2\alpha_1^2}} + n \text{, ~where~ } n\sim \mathcal{N}(0, \mathbf{\alpha}_2).
    \end{equation}
    We evaluate the performance gains using this surrogate function with the proposed AGAT training on the challenging CIFAR-10-C dataset.

    \subsubsection{Baselines:}
    Test-Time Training (TTT)~\cite{sun2020test} is a recent approach in which a classifier is trained only on source data, but the test sample is utilized to update the classifier during inference.
    Adversarial Logit Pairing (ALP)~\cite{kannan2018adversarial}, a technique for defending against adversarial attacks, and pixel-wise domain augmentation techniques MADA~\cite{qiao2020learning} and GUD~\cite{volpi2018generalizing} are also considered as baselines.
    We use ResNet-26~\cite{he2016deep} specially designed for CIFAR-10~\shortcite{russakovsky2015imagenet}, with group normalization~\cite{wu2018group} which is stable with different batch sizes.
    This acts as the naive classifier-only baseline (B).
    We also consider the classifier trained with an auxiliary self-supervised task of angle prediction~\cite{gidaris2018unsupervised} (B+SS).
    Our joint-training (JT) baseline is from TTT based on \cite{hendrycks2018benchmarking}.

    We compare three versions of our model: with additive noise only, with Gaussian filtering, and with a composition of Gaussian filter and noise. 
    Our models are trained for 150 epochs including pre-training epochs $N_{pre}\mathrm{=}100$, batch-size 128, and $M\mathrm{=}15$ update steps for adversarial augmentation.
    The number of augmented samples is $30\%$ of the original source data, and augmentation interval $N_{aug}$ is fixed at 2 epochs.
    For our model the coefficients in Equations~\ref{eq:loss_constraint}, and~\ref{eq:loss_distill} are: $\lambda_1=0.5, \lambda_2=0.5, \beta=0.25$.
    The learning rates $\eta, \mu$ for the classifier and adversarial augmentation are both 5e-5.

    \subsubsection{Results}
    \input{tables/cifar_by_category}
    In Table~\ref{tab:cifar_cat} we show the classification accuracies on CIFAR10-C.
    It can be seen that our method consistently outperforms all baselines overall, and also on three of the four categories of corruptions (weather, blur, and digital).
    It is interesting to note that the ALP performance on the Noise category is distinctly greater than all previous methods, potentially because it is designed to defend against projected gradient descent adversarial attacks~\cite{madry2017towards}.
    ALP uses a similar loss function as Equation~\ref{eq:loss_distill} to train the classifier, but still operates in pixel-space and does not perturb the attribute space.
    In Table~\ref{tab:cifar_cat} we also demonstrate that our models which uses only blur or only noise as surrogate are also better than previous state-of-the art. Note that the ``noise only'' model is in essence a pixel-level perturbation achieved by only perturbing along the variance parameter using AGAT training, and yet we see a significant boost in performance over all other pixel-level additive noise methods.
    Similarly, the ``blur only" model also gives performance boosts on weather and digital categories, further indicating the general applicability of our AGAT training approach.

%% file: tables/clevr_splits.tex
\begin{table}[t]
    \centering
    \small
    \begin{tabular*}{\linewidth}{@{}>{\raggedright}P{15mm}@{\extracolsep{\fill}}p{30mm}p{30mm}@{}}
        \toprule
        \textbf{Attribute} & \textbf{Train} & \textbf{Test} \\
        \toprule
        
        {{Size and Position}} & {(small, NW), (medium, NE), (large, SE)} & {(small, SE), (medium, NW), (large, SW)} \\
        \midrule
        {Material and Position} & {(metal, NW), (rubber, NW), (metal, NE), (rubber, NE)} & (rubber, SW), (metal, SW), (rubber, SE), (metal, SE)\\
        \bottomrule
    \end{tabular*}
    \caption{The train and test splits for our experiments with semantic object-level perturbations for CLEVR-Singles.}
    \label{tab:clevr_splits}
\end{table}

%% file: tables/clevr_results.tex
\begin{table}[t]
    \centering
    \begin{tabular*}{\linewidth}{@{}l@{\extracolsep{\fill}}ccc@{}}
        \toprule
        \textbf{Method} & \textbf{Source} & \textbf{Size+Pos.} & \textbf{Mat.+Pos.} \\ 
        \midrule
        B           & 99.81 & 89.92 & 59.90 \\
        GUD~\shortcite{volpi2018generalizing}         & 99.94 & 93.69 & 65.03 \\
        M-ADA~\shortcite{qiao2020learning}      & 99.96 & 94.52 & 65.50 \\
        Ours        & 99.97 & 95.22 & 69.49 \\ 
        \bottomrule
    \end{tabular*}
    \caption{Classification accuracy for color-classification on CLEVR-Singles. 
    Source and target sets are split on \textit{size+position} attribute for the third column, and \textit{Material+Position} for the fourth column.}
    \label{tab:clevr_results}
\end{table}

%% file: tables/mnist_rts.tex
\begin{table}[t]
    \centering
    \begin{tabular*}{\linewidth}{@{}l@{\extracolsep{\fill}}cccc@{}}
        \toprule
        \textbf{Method} & \textbf{R} & \textbf{T} & \textbf{S} & \textbf{RTS}\\
        \toprule
        B & 84.44 & 27.67 & 95.76 & 21.91 \\
        GUD~\shortcite{volpi2018generalizing} & 86.08 & 29.09 & 97.89 & 23.10 \\
        MADA~\shortcite{qiao2020learning} & 87.37 & 29.25 & \textbf{98.32} & {22.68} \\
        PS~\shortcite{wong2020learning} & \textbf{87.86} & 45.36 & 96.00 & 39.38 \\
        Ours ($T_{aug}\text{=}30\%$) & \underline{84.93} & \underline{\textbf{52.95}} & \underline{96.11} & \textbf{\underline{41.43}} \\
        \bottomrule
    \end{tabular*}
    \caption{Results on the MNIST-RTS robustness benchmark for rotation (R), translation (T), scaling (S), and random combination (RTS).}
    \label{tab:mnist_geom}
\end{table}

%% file: tables/mnist_ablation.tex
\begin{table}[t]
    \centering 
    \begin{tabular}{@{}cccccc@{}}
        \toprule
        $\mathbf{N_{aug}}$ & $\mathbf{T_{aug}}$ & \textbf{R} & \textbf{T} & \textbf{S} & \textbf{RTS} \\
        \toprule
        1 & 10 & 84.12 & 43.33 & 96.65 & 34.12 \\
         & 30 & 83.80 & 54.17 & 95.89 & 40.46 \\
         & 50 & 84.49 & 59.97 & 96.29 & 47.62 \\ 
         & 70 & 84.35 & 62.76 & 96.24 & 51.13 \\
        \midrule
        2 & 10 & 84.97 & 47.21 & 96.41 & 36.84 \\
         & 30 & 84.93 & 52.95 & 96.11 & 41.43 \\
         & 50 & 86.35 & 61.07 & 95.76 & 47.59 \\
         & 70 & 84.59 & 62.75 & 95.79 & 50.28\\
        \bottomrule
    \end{tabular}
    \caption{The effect of augmentation interval ($N_{aug}$) at different percentages of augmented samples 
    ($T_{aug}$). 
    }
    \label{}
\end{table}

\begin{table}[t]
    \centering 
    \begin{tabular}{@{}cccccc@{}}
        \toprule
        \textbf{Loss} & $\mathbf{T_{aug}}$ & \textbf{R} & \textbf{T} & \textbf{S} & \textbf{RTS} \\
        \toprule
        GT 
         & 10 & 85.41 & 29.48 & 96.74 & 23.57\\
         & 30 & 84.82 & 48.46 & 96.75 & 37.80\\
         & 50 & 84.17 & 52.44 & 95.86 & 41.82 \\
         & 70 & 84.07 & 55.14 & 95.70 & 44.20 \\
        \midrule
        GT~+~CR
         & 10 & 84.97 & 47.21 & 96.41 & 36.84 \\
         & 30 & 84.93 & 52.95 & 96.11 & 41.43 \\
         & 50 & 86.35 & 61.07 & 95.76 & 47.59 \\ 
         & 70 & 84.59 & 62.75 & 95.79 & 50.28 \\
        \bottomrule
    \end{tabular}
    \caption{
        The effect of classification loss function at different percentages of augmented samples. GT denotes the first term and CR is consistency regularization in Eq~\eqref{eq:loss_distill}.
    }
    \label{tab:mnist_ablation}
\end{table}

%% file: tables/cifar_by_category.tex
\begin{table}[t]
    \centering
    \begin{tabular*}{\linewidth}{@{}l cccccc@{}}
        \toprule 
        \textbf{Method} & \textbf{Src.} & \textbf{W} & \textbf{B} & \textbf{N} & \textbf{D} & \textbf{Avg.} \\
        \toprule 
        B       & 90.6 & 70.6 & 69.0 & 45.5 & 71.6 & 66.4\\
        B+SS    & 91.1 & 70.6 & 68.5 & 48.7 & 69.7 & 67.0\\
        GUD     & - & 71.7 & 59.2 & 30.5 & 64.7 & 58.3\\
        MADA    & - & 75.6 & 63.8 & 54.2 & 65.1 & 65.6\\
        JT      & 91.9 & 71.7 & 69.0 & 50.6 & 71.6 & 68.3\\
        ALP     & 83.5 & 60.9 & 74.7 & \textbf{75.4} & 68.5& 70.0\\
        TTT     & \textbf{92.1} & 73.7 & 71.3 & 54.2 & 73.4 & 70.5\\
        \midrule
        Ours (b. only) & \underline{91.3} & \underline{\textbf{78.4}} & \underline{\textbf{75.1}} & 49.3 & \underline{\textbf{75.4}} & 70.0\\
        Ours (n. only) & 90.3 & 75.0 & 73.3 & 62.4 & 73.1 & 71.3 \\
        Ours    & 89.3 & 77.8 & 74.1 & \underline{65.8} & 71.6 & \underline{\textbf{72.3}} \\
        \bottomrule
    \end{tabular*}
    \caption{Comparison of classification accuracies on CIFAR-10-C corruption categories (Weather, Blur, Noise, Digital). Our best scores are underlined; overall best are bold.}
    \label{tab:cifar_cat}
\end{table}

%% file: appendix_arxiv.tex
\section*{Appendix}
This appendix contains
further details about the creation of the CLEVR-Singles dataset, 
additional sample images generated by AGAT, 
model details, 
and fine-grained results on the CIFAR-10-C benchmark.

\section{CLEVR-Singles Dataset Creation}
The CLEVR-Singles dataset has been created to allow a diagnostic setup with discrete and controllable set of attributes for our robustness experiments for semantic object-level perturbations.
In this section we delineate the dataset creation process and provide more examples from the dataset.

We use Blender~\cite{blender} to render images in CLEVR-Singles.
Every scene contains a single object, have a randomly chosen size, shape, material, position, and color.
Our code for rendering these images is a modification of the data creation process for the CLEVR dataset~\cite{johnson2017clevr}.
The choices for each property of the object are shown in Table~\ref{tab:clevr_properties}.

\begin{table}[H]
    \centering
    \small
    \begin{tabular*}{\linewidth}{@{}ll@{}}
        \toprule
        \textbf{Property} &  \textbf{Choices}\\
        \midrule
        Color   & gray, red, blue, green, brown, purple, cyan, yellow \\
        Size    & large (1.2), medium (0.9), small (0.6) \\
        Shape   & Cube, Sphere, Cylinder, Pyramid \\ 
        Material& Rubber, Metal \\
        Position & NW, NE, SW, SE \\
        Rotation & $\theta \in [-180, 180]$ \\
        \bottomrule
    \end{tabular*}
    \caption{List of properties that can be assigned to an object to render the images in CLEVR-Singles.}
    \label{tab:clevr_properties}
\end{table} 

\noindent Size refers to the height or diameter of the shapes which are symmetric along $x$ and $y$ axes.
In terms of 3D coordinates, the objects are placed at the position $(X, Y, Z)$ such that: $$X \in [-3, 3], Y \in [-3, 3], Z = 0.5*size.$$
For our experiments, we define position to be one of the four quadrants of the $xy$ plane and denote these by NW, NE, SW, SE, which allows us to easily split the dataset for robustness experiments.

The positions of the lighting and camera and are slightly jittered to introduce variation.
The objects are rotated at a random angle while placing.
We generate images of size $256\times256$ using a tile size of $128$ for rendering.
The CLEVR-Singles dataset contains $50000$ images for training, $10000$ for validation, and $10000$ test samples.
Each image takes approximately 5 seconds to render using a single GPU, and can also be rendered on CPU at 8 seconds per image.
Along with the images, scene graphs are also generated which contain labels for each attribute, color, and other properties.

\begin{figure*}[t]
    \centering
    \includegraphics[width=\linewidth]{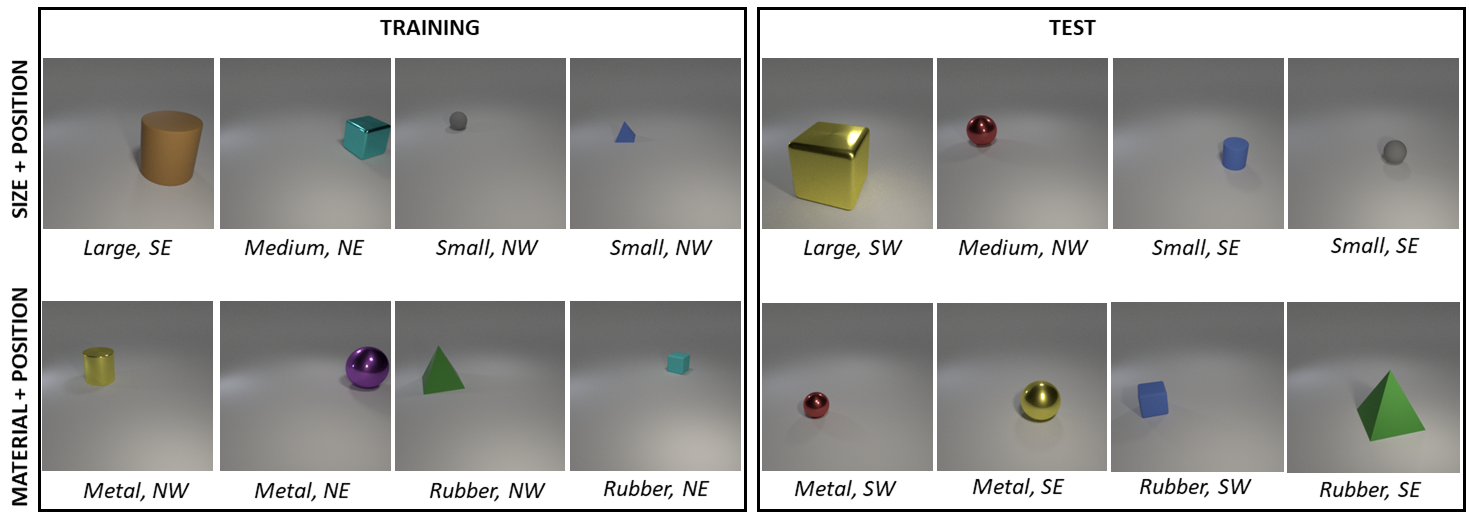}
    \caption{Sample images from the training and test splits for robustness experiments on the CLEVR-Singles dataset. The first row shows the train-test split on the attributes \textit{size+position}, and the second row for \textit{material+position}.}
    \label{fig:clevr_illus_supp}
\end{figure*}

For our experimental setup for robustness experiments, we split the dataset into training and test splits according to specific attributes in Table~\ref{tab:clevr_splits_supp}.

\begin{table}[H]
    \centering
    \small
    \begin{tabular*}{\linewidth}{@{}>{\raggedright}P{15mm}@{\extracolsep{\fill}}p{30mm}p{30mm}@{}}
        \toprule
        \thead{Attribute} & \thead{Train} & \thead{Test} \\
        \toprule
        
        {{Size and Position}} & {(small, NW), (medium, NE), (large, SE)} & {(small, SE), (medium, NW), (large, SW)} \\
        \midrule
        {Material and Position} & {(metal, NW), (rubber, NW), (metal, NE), (rubber, NE)} & (rubber, SW), (metal, SW), (rubber, SE), (metal, SE)\\
        \bottomrule
    \end{tabular*}
    \caption{The train and test splits for our experiments with semantic object-level perturbations for CLEVR-Singles.}
    \label{tab:clevr_splits_supp}
\end{table}

Illustrative examples are shown in Figure~\ref{fig:clevr_illus_supp} with the first row showing samples from training and test splits for \textit{size+position}, and the second row for \textit{material+position}.
The data-generation code and train-test splits for robustness experiments will be made publicly available upon publication.

\section{AGAT-Generated Images}
 We provide additional samples generated by attribute-guided adversarial training algorithm (AGAT) for each of our experiments.

    \subsubsection{Semantic Object-Level Perturbations:}

\begin{figure}[t]
        \centering
        \includegraphics[width=\linewidth]{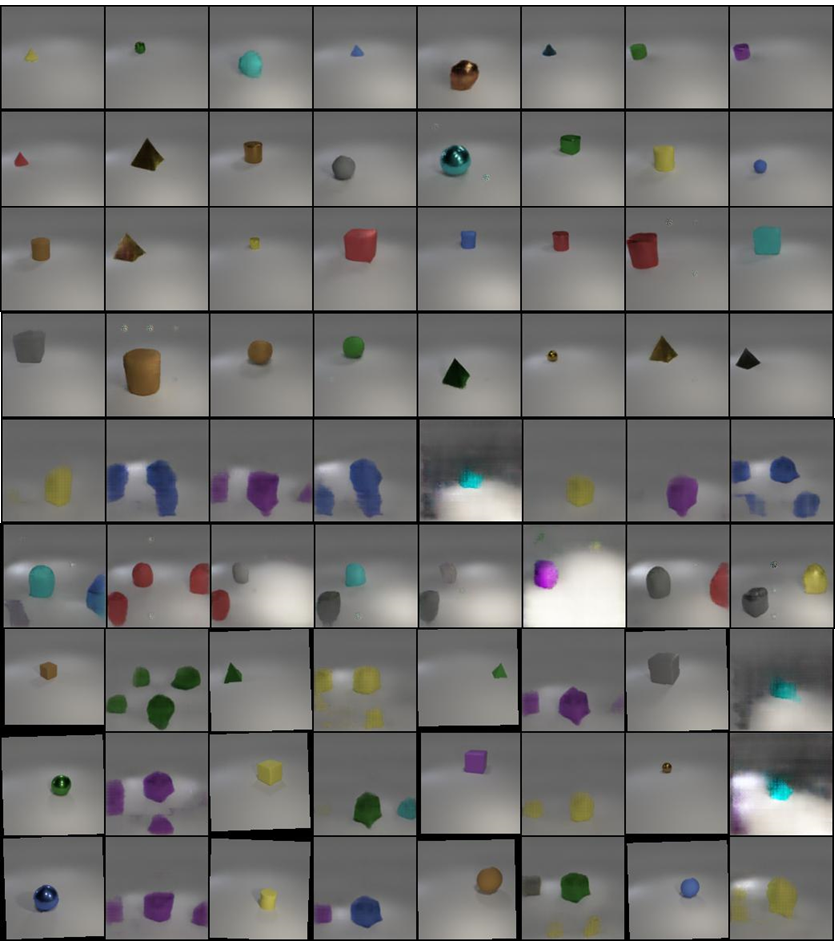}
        \caption{
        Examples of images generated by AGAT for the CLEVR-Singles dataset. The AttGAN generator is able to explore the attribute space and generate images with different attributes, as well as novel scenes such as those containing multiple objects.
        }
        \label{fig:supp_gen_clevr}
    \end{figure}

    Figure~\ref{fig:supp_gen_clevr} shows the images generated by AttGAN~\cite{he2019attgan} during AGAT.
    It can be seen that AttGAN is able to explore various sizes, materials during the augmentation, as well generate novel scenes with multiple objects which are not present in the training dataset.

    \subsubsection{Geometric Transformations:} 

\begin{figure}[t]
        \centering
        \includegraphics[width=\linewidth]{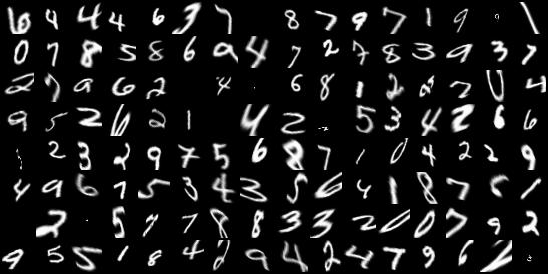}
        \caption{
        Examples of images generated by AGAT for MNIST. AGAT is able to explore the attribute space and generate images with different rotation, translations, and skews. 
        }
        \label{fig:supp_gen_mnist}
    \end{figure}

    Figure~\ref{fig:supp_gen_mnist} shows the images generated by STN~\cite{jaderberg2015spatial} during AGAT training.
    It can be seen that rotations, translations, skew, and combinations of these are generated.
    
    \subsubsection{Common Image Corruptions:} 
\begin{figure}[t]
    \centering
    \includegraphics[width=\linewidth]{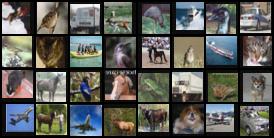}
    \caption{
    Examples of images generated by AGAT for the CIFAR-10 dataset. AGAT is able to explore the attribute space and generate images having varying degrees of noise and blur, which helps improve robustness on all 15 categories of corruptions as seen in Table~\ref{tab:cifar_nat}.        
    }
    \label{fig:supp_gen_cifar}
\end{figure}

    Figure~\ref{fig:supp_gen_cifar} shows the images during AGAT training on the CIFAR-10 dataset using additive Gaussian noise and blur as the surrogate function.

\input{tables/cifar}

\section{Model Architectures}
We use the same classifier for baselines as well as our model for fair comparison.
For experiments on CLEVR-Singles, we use a color-classification neueral network with 4 convolutional layers with stride $2$ and kernel-size $3$, followed by 2 fully-connected layers. 
For experiments on MNIST, we use a classifier with two convolutional layers with kernel-size $5$ and stride $1$ and 3 fully-connected layers.
For CIFAR-10-C experiments we use the classifier architecture used by \cite{sun2020test} for fair comparison. The classifier is a ResNet~\cite{he2016deep} constructed specially for CIFAR-10 images and has 26 layers, group norm~\cite{wu2018group} with 8 layers.

\section{Additional CIFAR-10-C Results}
In Table~\ref{tab:cifar_nat} we provide detailed comparison of our models with the baselines on the CIFAR-10-C benchmark for each of the 15 categories of corruptions.
Our method shows consistent improvements across all 15 categories when compares with the classifier-only baselines (``B" and ``B+SS") as well as the current state-of-the-art model Test-Time Training (TTT)~\cite{sun2020test}.
M-ADA~\cite{qiao2020learning} is the best performing for ``Snow" and ``Brightness" images, but does not show consistent improvements on other categories.
Similarly, ALP~\cite{kannan2018adversarial} is the best on all four noise categories, but suffers significantly on ``Fog" ($35.2\%$) and ``Contrast" ($26.4\%$) compared to all other baselines.
Table~\ref{tab:cifar_nat} also shows the variances of each approach. 
Our approach (models using noise only, or both noise and blur) have a standard deviation lower than all other approaches.



%% file: tables/cifar.tex
\begin{table*}[t]
        \centering
        \small
        \resizebox{\linewidth}{!}{
        \begin{tabular}{@{}l c ccc  ccccc  cccc  ccccccc  cc@{}}
            \toprule
            \textbf{Method} & \textbf{SRC}
                & Fog & Snow & Frst
                & Zoom & Defo & Glas 
                & Motn
                & Shot & Impl & Gaus
                & JPEG & Pixl 
                & Elas & Brit 
                & Cont
                & \textbf{Avg} $\pm$ \textbf{Std}\\
            \toprule
            B   & 90.6
                & 73.1 & 74.5 & 63.3
                & 73.6 & 74.9 & 49.8 
                & 77.6
                & 49.1 & 42.6 & 44.9
                & 71.3 & 54.9 
                & 73.4 & 87.2 
                & 71.2
                & 66.4 $\pm$ 13.2\\
            B+SS   & 91.1 
                & 71.9 & 74.4 & 65.6 
                & 73.7 & 76.3 & 48.3 
                & 75.7 
                & 52.8 & 43.9 & 49.5 
                & 70.2 & 44.2 
                & 72.6 & 86.5 
                & 75.0 
                & 67.0 $\pm$ 13.3\\
            GUD & {-} 
                & {68.3} & {76.8} & {70.0} 
                & {62.9} & {56.4} & {53.5} 
                & {63.9} 
                & {36.9} & {22.3} & {32.4} 
                & {74.2} & {53.3} 
                & {74.6} & {89.9} 
                & {31.6} 
                & {58.3 $\pm$ 18.9}\\
            MADA & {-} 
                & {69.4} & \textbf{80.6} & {76.7} 
                & {68.0} & {61.2} & {61.6} 
                & {64.2} 
                & {60.6} & {45.2} & {56.9} 
                & {77.1} & {52.3} 
                & {75.6} & \textbf{90.8} 
                & {29.7} 
                & {65.6 $\pm$ 14.7} \\
            JT & {91.9} 
                & 72.5 & 75 & 67.5 
                & 73.6 & 75.8 & 51.5 
                & 75.2 
                & 54.7 & 46.6 & 50.6 
                & 71.3 & 48.4 
                & 76 & 87.4 
                & 74.7 
                & 68.3 $\pm$ 12.3\\
            ALP & 83.5 
                & 35.2 & 74.8 & 72.8 
                & 76.9 & 75 & \textbf{74.4} 
                & 72.6 
                & \textbf{77.1} & \textbf{71.7} & \textbf{77.3} 
                & \textbf{81.1} & \textbf{79.8} 
                & 77.0 & 78.3 
                & 26.4 
                & 70.0 $\pm$ 15.7\\
            TTT & \textbf{92.1 }
                & 74.9 & 76.1 & 70.0 
                & 76.1 & 78.2 & 53.9 
                & 77.0 
                & 58.2 & 50.0 & 54.4 
                & 72.8 & 52.8 
                & \textbf{77.4} & 87.8 
                & 76.1 
                & 70.5 $\pm$ 11.4\\
            \midrule
            Ours (blur only) & \underline{91.3}
                    & 77.0 & \underline{76.5} & 71.2 & \textbf{\underline{79.7}} & 83.8 & 57.9 & 79.0 & 51.5 & 49.1 & 47.4 & 70.5 & 52.9 & \underline{75.0} & \underline{87.0} & 78.7
                & 70.6 $\pm$ 13.1\\
            Ours (noise only) & 90.3
                & 76.1 & 75.6 & 73.3 
                & 77.2 & 80.3 & 57.6 & 78.0
                & 64.9 & 56.3 & 65.9
                & \underline{73.4} & \underline{57.2} & 72.8 & 85.8 & 76.3  
                & 71.3 $\pm$ 8.7\\
            Ours (both) & 89.3 
                & \underline{\textbf{79.0}} & 76.2 & \underline{\textbf{79.2}} 
                & 76.4 & \underline{\textbf{80.3}} & \underline{61.4} & \underline{\textbf{78.2}} 
                & \underline{70.3} & \underline{59.5} & \underline{67.7} 
                & 69.5 & 48.6 & 72.8 & 85.4 & \textbf{\underline{81.5}} 
                & \underline{\textbf{72.3}}$\pm$ 9.5\\
            \bottomrule
        \end{tabular}
        }
        \caption{Comparison of classification accuracies on CIFAR-10-C for each type of corruption. \underline{Underlined} are our best scores, and \textbf{bold} are the overall best.}
        \label{tab:cifar_nat}
    \end{table*}

%% file: Formatting-Instructions-LaTeX-2021.bbl
\begin{thebibliography}{34}
\providecommand{\natexlab}[1]{#1}
\providecommand{\url}[1]{\texttt{#1}}
\providecommand{\urlprefix}{URL }
\expandafter\ifx\csname urlstyle\endcsname\relax
  \providecommand{\doi}[1]{doi:\discretionary{}{}{}#1}\else
  \providecommand{\doi}{doi:\discretionary{}{}{}\begingroup
  \urlstyle{rm}\Url}\fi

\bibitem[{Blender Online~Community(2018)}]{blender}
Blender Online~Community, A. 2018.
\newblock \emph{Blender - a 3D modelling and rendering package}.
\newblock Blender Foundation, Stichting Blender Foundation, Amsterdam.
\newblock \urlprefix\url{http://www.blender.org}.

\bibitem[{Bolukbasi et~al.(2016)Bolukbasi, Chang, Zou, Saligrama, and
  Kalai}]{bolukbasi2016man}
Bolukbasi, T.; Chang, K.-W.; Zou, J.~Y.; Saligrama, V.; and Kalai, A.~T. 2016.
\newblock Man is to computer programmer as woman is to homemaker? debiasing
  word embeddings.
\newblock In \emph{Advances in neural information processing systems},
  4349--4357.

\bibitem[{Bulusu et~al.(2020)Bulusu, Kailkhura, Li, Varshney, and
  Song}]{bulusu2020anomalous}
Bulusu, S.; Kailkhura, B.; Li, B.; Varshney, P.~K.; and Song, D. 2020.
\newblock Anomalous Example Detection in Deep Learning: A Survey.
\newblock \emph{IEEE Access} 8: 132330--132347.

\bibitem[{Cohen and Welling(2014)}]{cohen2014transformation}
Cohen, T.~S.; and Welling, M. 2014.
\newblock Transformation properties of learned visual representations.
\newblock \emph{arXiv preprint arXiv:1412.7659} .

\bibitem[{Gidaris, Singh, and Komodakis(2018)}]{gidaris2018unsupervised}
Gidaris, S.; Singh, P.; and Komodakis, N. 2018.
\newblock Unsupervised Representation Learning by Predicting Image Rotations.
\newblock In \emph{International Conference on Learning Representations}.

\bibitem[{Goodfellow, Shlens, and Szegedy(2014)}]{goodfellow2014explaining}
Goodfellow, I.~J.; Shlens, J.; and Szegedy, C. 2014.
\newblock Explaining and harnessing adversarial examples.
\newblock \emph{arXiv preprint arXiv:1412.6572} .

\bibitem[{He et~al.(2016)He, Zhang, Ren, and Sun}]{he2016deep}
He, K.; Zhang, X.; Ren, S.; and Sun, J. 2016.
\newblock Deep residual learning for image recognition.
\newblock In \emph{Proceedings of the IEEE conference on computer vision and
  pattern recognition}, 770--778.

\bibitem[{He et~al.(2019)He, Zuo, Kan, Shan, and Chen}]{he2019attgan}
He, Z.; Zuo, W.; Kan, M.; Shan, S.; and Chen, X. 2019.
\newblock Attgan: Facial attribute editing by only changing what you want.
\newblock \emph{IEEE Transactions on Image Processing} 28(11): 5464--5478.

\bibitem[{Hendricks et~al.(2018)Hendricks, Burns, Saenko, Darrell, and
  Rohrbach}]{hendricks2018women}
Hendricks, L.~A.; Burns, K.; Saenko, K.; Darrell, T.; and Rohrbach, A. 2018.
\newblock Women also snowboard: Overcoming bias in captioning models.
\newblock In \emph{European Conference on Computer Vision}, 793--811. Springer.

\bibitem[{Hendrycks and Dietterich(2018)}]{hendrycks2018benchmarking}
Hendrycks, D.; and Dietterich, T. 2018.
\newblock Benchmarking Neural Network Robustness to Common Corruptions and
  Perturbations.
\newblock In \emph{International Conference on Learning Representations}.

\bibitem[{Isola et~al.(2017)Isola, Zhu, Zhou, and Efros}]{isola2017image}
Isola, P.; Zhu, J.-Y.; Zhou, T.; and Efros, A.~A. 2017.
\newblock Image-to-image translation with conditional adversarial networks.
\newblock In \emph{Proceedings of the IEEE conference on computer vision and
  pattern recognition}, 1125--1134.

\bibitem[{Jaderberg et~al.(2015)Jaderberg, Simonyan, Zisserman
  et~al.}]{jaderberg2015spatial}
Jaderberg, M.; Simonyan, K.; Zisserman, A.; et~al. 2015.
\newblock Spatial transformer networks.
\newblock In \emph{Advances in neural information processing systems},
  2017--2025.

\bibitem[{Johnson et~al.(2017)Johnson, Hariharan, van~der Maaten, Fei-Fei,
  Lawrence~Zitnick, and Girshick}]{johnson2017clevr}
Johnson, J.; Hariharan, B.; van~der Maaten, L.; Fei-Fei, L.; Lawrence~Zitnick,
  C.; and Girshick, R. 2017.
\newblock Clevr: A diagnostic dataset for compositional language and elementary
  visual reasoning.
\newblock In \emph{Proceedings of the IEEE Conference on Computer Vision and
  Pattern Recognition}, 2901--2910.

\bibitem[{Joshi et~al.(2019)Joshi, Mukherjee, Sarkar, and
  Hegde}]{joshi2019semantic}
Joshi, A.; Mukherjee, A.; Sarkar, S.; and Hegde, C. 2019.
\newblock Semantic adversarial attacks: Parametric transformations that fool
  deep classifiers.
\newblock In \emph{Proceedings of the IEEE International Conference on Computer
  Vision}, 4773--4783.

\bibitem[{Kannan, Kurakin, and Goodfellow(2018)}]{kannan2018adversarial}
Kannan, H.; Kurakin, A.; and Goodfellow, I. 2018.
\newblock Adversarial logit pairing.
\newblock \emph{arXiv preprint arXiv:1803.06373} .

\bibitem[{Karras, Laine, and Aila(2019)}]{karras2019style}
Karras, T.; Laine, S.; and Aila, T. 2019.
\newblock A style-based generator architecture for generative adversarial
  networks.
\newblock In \emph{Proceedings of the IEEE conference on computer vision and
  pattern recognition}, 4401--4410.

\bibitem[{Krizhevsky(2009)}]{krizhevsky2009learning}
Krizhevsky, A. 2009.
\newblock Learning Multiple Layers of Features from Tiny Images.
\newblock \emph{Master's thesis, University of Toronto} .

\bibitem[{LeCun et~al.(1998)LeCun, Bottou, Bengio, and
  Haffner}]{lecun1998mnist}
LeCun, Y.; Bottou, L.; Bengio, Y.; and Haffner, P. 1998.
\newblock Gradient-based learning applied to document recognition.
\newblock \emph{Proceedings of the IEEE} 86(11): 2278--2324.

\bibitem[{Liu et~al.(2018)Liu, Tao, Li, Nowrouzezahrai, and
  Jacobson}]{liu2018beyond}
Liu, H.-T.~D.; Tao, M.; Li, C.-L.; Nowrouzezahrai, D.; and Jacobson, A. 2018.
\newblock Beyond pixel norm-balls: Parametric adversaries using an analytically
  differentiable renderer.
\newblock \emph{arXiv preprint arXiv:1808.02651} .

\bibitem[{Madry et~al.(2017)Madry, Makelov, Schmidt, Tsipras, and
  Vladu}]{madry2017towards}
Madry, A.; Makelov, A.; Schmidt, L.; Tsipras, D.; and Vladu, A. 2017.
\newblock Towards deep learning models resistant to adversarial attacks.
\newblock \emph{arXiv preprint arXiv:1706.06083} .

\bibitem[{Madry et~al.(2018)Madry, Makelov, Schmidt, Tsipras, and
  Vladu}]{madry2018towards}
Madry, A.; Makelov, A.; Schmidt, L.; Tsipras, D.; and Vladu, A. 2018.
\newblock Towards Deep Learning Models Resistant to Adversarial Attacks.
\newblock In \emph{International Conference on Learning Representations}.

\bibitem[{Mirza and Osindero(2014)}]{mirza2014conditional}
Mirza, M.; and Osindero, S. 2014.
\newblock Conditional generative adversarial nets.
\newblock \emph{arXiv preprint arXiv:1411.1784} .

\bibitem[{Qiao, Zhao, and Peng(2020)}]{qiao2020learning}
Qiao, F.; Zhao, L.; and Peng, X. 2020.
\newblock Learning to learn single domain generalization.
\newblock In \emph{Proceedings of the IEEE/CVF Conference on Computer Vision
  and Pattern Recognition}, 12556--12565.

\bibitem[{Raghunathan, Steinhardt, and Liang(2018)}]{raghunathan2018certified}
Raghunathan, A.; Steinhardt, J.; and Liang, P. 2018.
\newblock Certified Defenses against Adversarial Examples.
\newblock In \emph{International Conference on Learning Representations}.

\bibitem[{Recht et~al.(2018)Recht, Roelofs, Schmidt, and
  Shankar}]{recht2018cifar}
Recht, B.; Roelofs, R.; Schmidt, L.; and Shankar, V. 2018.
\newblock Do cifar-10 classifiers generalize to cifar-10?
\newblock \emph{arXiv preprint arXiv:1806.00451} .

\bibitem[{Russakovsky et~al.(2015)Russakovsky, Deng, Su, Krause, Satheesh, Ma,
  Huang, Karpathy, Khosla, Bernstein et~al.}]{russakovsky2015imagenet}
Russakovsky, O.; Deng, J.; Su, H.; Krause, J.; Satheesh, S.; Ma, S.; Huang, Z.;
  Karpathy, A.; Khosla, A.; Bernstein, M.; et~al. 2015.
\newblock Imagenet large scale visual recognition challenge.
\newblock \emph{International journal of computer vision} 115(3).

\bibitem[{Sinha, Namkoong, and Duchi(2018)}]{sinha2018certifying}
Sinha, A.; Namkoong, H.; and Duchi, J. 2018.
\newblock Certifying Some Distributional Robustness with Principled Adversarial
  Training.
\newblock In \emph{International Conference on Learning Representations}.

\bibitem[{Sun et~al.(2020)Sun, Wang, Liu, Miller, Efros, and
  Hardt}]{sun2020test}
Sun, Y.; Wang, X.; Liu, Z.; Miller, J.; Efros, A.~A.; and Hardt, M. 2020.
\newblock Test-time training with self-supervision for generalization under
  distribution shifts.
\newblock In \emph{International Conference on Machine Learning (ICML)}.

\bibitem[{Volpi et~al.(2018)Volpi, Namkoong, Sener, Duchi, Murino, and
  Savarese}]{volpi2018generalizing}
Volpi, R.; Namkoong, H.; Sener, O.; Duchi, J.~C.; Murino, V.; and Savarese, S.
  2018.
\newblock Generalizing to unseen domains via adversarial data augmentation.
\newblock In \emph{Advances in neural information processing systems},
  5334--5344.

\bibitem[{Wong and Kolter(2020)}]{wong2020learning}
Wong, E.; and Kolter, J.~Z. 2020.
\newblock Learning perturbation sets for robust machine learning.
\newblock \emph{arXiv preprint arXiv:2007.08450} .

\bibitem[{Wu and He(2018)}]{wu2018group}
Wu, Y.; and He, K. 2018.
\newblock Group normalization.
\newblock In \emph{Proceedings of the European conference on computer vision
  (ECCV)}, 3--19.

\bibitem[{Xiao et~al.(2020)Xiao, Engstrom, Ilyas, and Madry}]{xiao2020noise}
Xiao, K.; Engstrom, L.; Ilyas, A.; and Madry, A. 2020.
\newblock Noise or Signal: The Role of Image Backgrounds in Object Recognition.
\newblock \emph{arXiv preprint arXiv:2006.09994} .

\bibitem[{Zhang et~al.(2017)Zhang, Xu, Li, Zhang, Wang, Huang, and
  Metaxas}]{zhang2017stackgan}
Zhang, H.; Xu, T.; Li, H.; Zhang, S.; Wang, X.; Huang, X.; and Metaxas, D.~N.
  2017.
\newblock Stackgan: Text to photo-realistic image synthesis with stacked
  generative adversarial networks.
\newblock In \emph{Proceedings of the IEEE international conference on computer
  vision}, 5907--5915.

\bibitem[{Zhao et~al.(2017)Zhao, Wang, Yatskar, Ordonez, and
  Chang}]{zhao2017men}
Zhao, J.; Wang, T.; Yatskar, M.; Ordonez, V.; and Chang, K.-W. 2017.
\newblock Men Also Like Shopping: Reducing Gender Bias Amplification using
  Corpus-level Constraints.
\newblock In \emph{Proceedings of the 2017 Conference on Empirical Methods in
  Natural Language Processing}, 2979--2989.

\end{thebibliography}
